\documentclass[10pt, a4paper]{article}
\usepackage{lrec2022} 
\usepackage{multibib}
\newcites{languageresource}{Language Resources}
\usepackage{graphicx}
\usepackage{tabularx}
\usepackage{soul}
\usepackage{titlesec}
\titleformat{\section}{\normalfont\large\bfseries\center}{\thesection.}{1em}{}
\titleformat{\subsection}{\normalfont\SmallTitleFont\bfseries\raggedright}{\thesubsection.}{1em}{}
\titleformat{\subsubsection}{\normalfont\normalsize\bfseries\raggedright}{\thesubsubsection.}{1em}{}
\renewcommand\thesection{\arabic{section}}
\renewcommand\thesubsection{\thesection.\arabic{subsection}}
\renewcommand\thesubsubsection{\thesubsection.\arabic{subsubsection}}

\usepackage{epstopdf}
\usepackage[utf8]{inputenc}

\usepackage[hidelinks]{hyperref}
\usepackage{xstring}

\usepackage{color}

\title{Combining State-of-the-Art Models with Maximal Marginal Relevance for Few-Shot and Zero-Shot Multi-Document Summarization}

\name{
    David Adams,\textsuperscript{\rm 1} Gandharv Suri,\textsuperscript{\rm 2} Yllias Chali\textsuperscript{\rm 1}
}

\address{
    david.adams@uleth.ca, gandharv.suri@iiitb.ac.in, yllias.chali@uleth.ca \\
    \textsuperscript{\rm 1}University of Lethbridge\\
    4401 University Drive\\
    Lethbridge, Alberta, Canada, T1K 3M4\\
    \textsuperscript{\rm 2}International Institute of Information Technology Bangalore\\
    26/C, Hosur Rd, Electronics City Phase 1\\
    Electronic City, Bengaluru, Karnataka, India, 560100
}

\abstract{
In Natural Language Processing, multi-document summarization (MDS) poses many challenges to researchers above those posed by single-document summarization (SDS). These challenges include the increased search space and greater potential for the inclusion of redundant information. While advancements in deep learning approaches have led to the development of several advanced language models capable of summarization, the variety of training data specific to the problem of MDS remains relatively limited. Therefore, MDS approaches which require little to no pretraining, known as few-shot or zero-shot applications, respectively, could be beneficial additions to the current set of tools available in summarization. To explore one possible approach, we devise a strategy for combining state-of-the-art models’ outputs using maximal marginal relevance (MMR) with a focus on query relevance rather than document diversity. Our MMR-based approach shows improvement over some aspects of the current state-of-the-art results in both few-shot and zero-shot MDS applications while maintaining a state-of-the-art standard of output by all available metrics.
 \\ \newline \Keywords{multidocument, summarization, natural language processing} }

\begin{document}

\maketitleabstract

\section{Introduction}
There have been consistent advancements in automatic text summarization due to its importance and relevance in modern natural language processing (NLP) applications. The task of single document summarization (SDS) remains a challenge for which various extractive and abstractive approaches have been researched. Further challenges arise in the task of multi-document summarization (MDS), which is the generation of a summary from a cluster of related documents as opposed to a single source document, and introduced additional challenges due to the increased search space and increased potential for redundancy. While various summarization models may be applicable to MDS document clusters as flattened documents, training data and pretrained models specific to MDS remain relatively limited, increasing the usefulness of few-shot and zero-shot approaches. In a few-shot application, the model or system being utilized has limited prior information regarding the target data. In this context, few-shot summarization involves the use of a system which is only partially pretrained on the corpus being summarized. In zero-shot applications, there is no prior knowledge of the target data and the system is not pretrained on the corpus at all.

Promising advancements in extractive summarization have been made, such as the use of text-matching \cite{zhong-etal-2020-extractive}. There is also a growing interest in the use of reinforcement-learning (RL) for text summarization, particularly for extractive summarization approaches. These approaches include the incorporation of simple embedding features in a RL summarization approach \cite{lee-lee-2017-automatic} and the use of RL for sentence-ranking \cite{narayan-etal-2018-ranking}. Additionally, some approaches to extractive summarization involve the use of simple statistical methods for the improved summaries, such as the various statistical models Daiya and Singh \shortcite{daiya-singh-2018-using} combined with various semantic models for MDS. Another of these statistical methods being utilized for summarization is the use of the maximal marginal relevance (MMR) algorithm, which Mao et al. \shortcite{mao-etal-2020-multi} incorporate into a reinforcement learning approach for MDS.

While extractive approaches remain effective in producing accurate, low-cost summaries from the source documents themselves, abstractive methods allow for more comprehensive summaries using novel terms from outside the source documents. Recent advancements in both extractive and abstractive text summarization methods have involved the development and use of innovative language models based on various architectures. While neural networks remain effective summarization model architectures with potential for advancements \cite{see-etal-2017-get,khatri2018abstractive,cohan-etal-2018-discourse}, there is also a substantial increase in summarization approaches using models based on the transformer architecture proposed by Vaswani \shortcite{vaswani2017attention}. Recent state-of-the-art models based on the transformer architecture include BERT \cite{devlin-etal-2019-bert}, RoBERTa \cite{liu2019roberta}, BART \cite{lewis-etal-2020-bart}, UniLM \cite{dong2019unified}, GPT-2 \cite{Radford2019LanguageMA}, PEGASUS \cite{zhang2020pegasus}, LED \cite{beltagy2020longformer}, ALBERT \cite{lan2020albert}, T5 \cite{raffel2020exploring}, ProphetNet \cite{qi-etal-2020-prophetnet}, XLNet \cite{yang2020xlnet}, and TED \cite{yang-etal-2020-ted}. 

State-of-the-art summarization models, including those which are not pretrained on multi-document datasets, can produce promising MDS results. However, the potential for oversized or redundant input data remains. It seems possible that if a summarization approach could utilize the diversity of outputs produced by pretrained models while minimizing their redundancy, the combined outputs could represent an improvement in the task of MDS, particularly in few-shot and zero-shot applications. To address the models' limitations with respect to few-shot and zero-shot MDS, we propose an abstractive-extractive MDS approach which combines state-of-the-art model outputs in a manner which improves state-of-the-art MDS performance. To achieve this combination of outputs, we employ the MMR algorithm with a strong bias towards MMR query relevance among model-generated outputs. We test this approach in both a few-shot and zero-shot context.

\section{Method}

Our framework makes use of the output of multiple pretrained models. Some of these models are pretrained on the MDS dataset being summarized, while some of the models are pretrained on a SDS dataset. Training all models on the MDS dataset would be expensive and unrelated to the goals of our approach, which we intend to prove effective without the need for optimal pretraining conditions. For any model which is trained on the MDS dataset being summarized, the source document cluster is used as a single-document input for the model. This is the input method for which these models were pretrained. For every model including those trained on the MDS dataset, each document cluster is split into single documents, for which a summary is produced. Preliminary research indicates that models which are not pretrained on the given MDS dataset produce more accurate summaries on the single documents individually than on the combined document cluster.

\subsection{MMR} MMR is a measure of relevance which is dependent on the amount of new information which is provided by each input document \cite{101145/290941291025}. The formula for MMR is as follows:
\begin{equation}
\resizebox{8cm}{!} 
{
	$ Arg \; \max_{D_{i \notin S}} [\lambda (Sim1(D_{i},Q)) - (1 - \lambda) \; (\max_{D_{j \in S}} Sim2(D_{i},D_{j}))] $
}
\end{equation}

In the MMR formula, the lambda constant $\lambda$ is a number between 0 and 1 which determines the degree to which the calculation prioritizes relevance or diversity. \textit{S} is the set of documents which have already been selected, \textit{$D_i$} is the given candidate document or sentence which is not selected, \textit{$D_j$} is the given previously-selected document to which the candidate document's similarity is compared, and \textit{Q} is the query document to which the relevance of each candidate document is computed. Given a desired number of documents to select, the MMR calculation iterates through the unselected documents and selects the desired number of documents that are the most relevant or the most diverse, depending on the lambda constant. \textit{Sim1} and \textit{Sim2} are similarity measurements between documents. However, given the $(1 - \lambda)$ factor preceding \textit{Sim2}, the right-hand side of the calculation effectively becomes a maximization of diversity rather than similarity. A higher lambda constant increases relevance to the query, while a lower lambda constant increases diversity among the selected documents. If only one document is passed through the algorithm, only the relevance portion of the algorithm is used, as is also the case with the first document selection when running the MMR algorithm. We intentionally choose a sentence from our best-performing model as the first selected document.

\subsection{Approach Overview} 
\begin{figure}[ht]
	\centering
	\includegraphics[width=\linewidth]{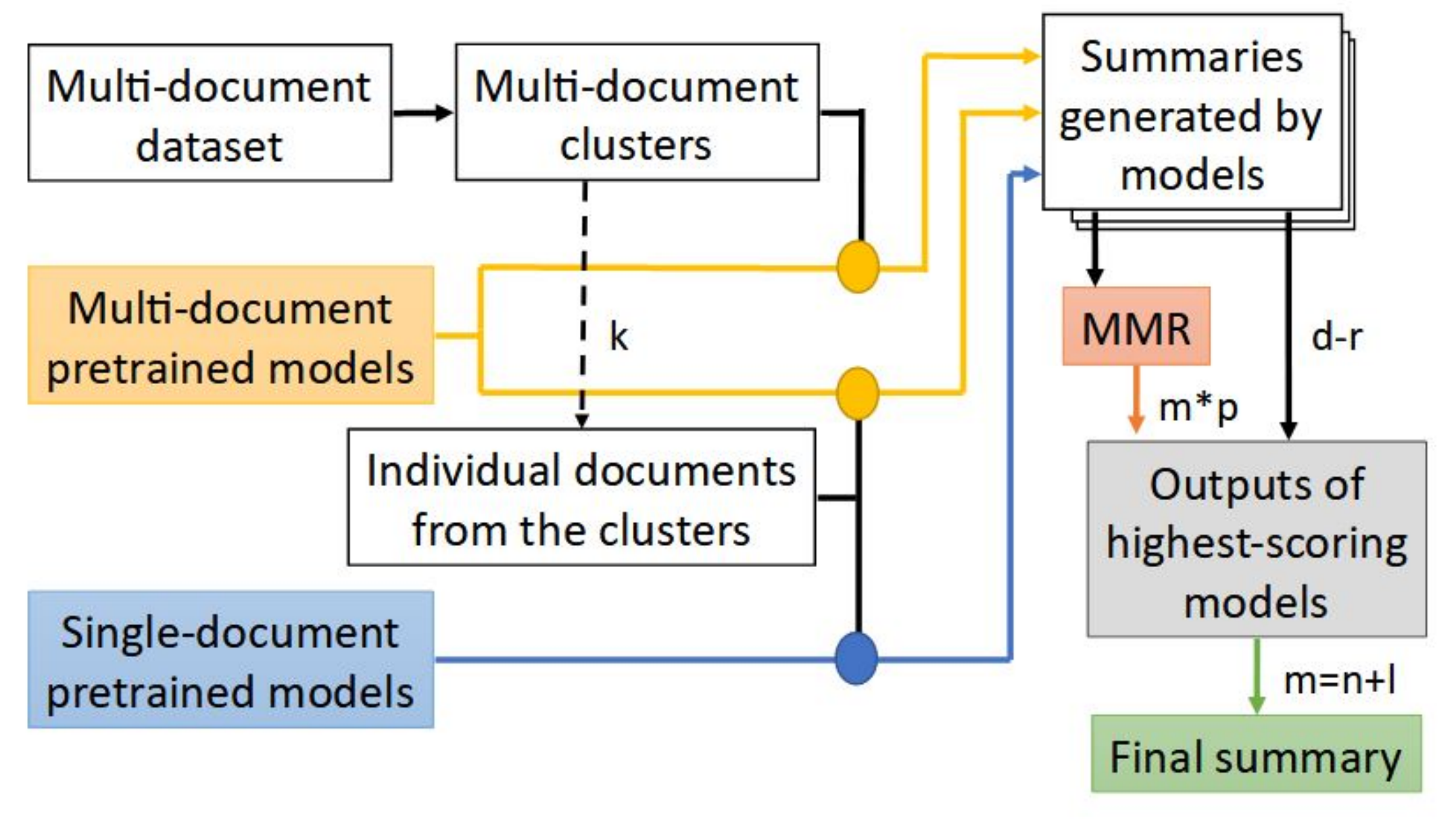}
	\caption{Overview of our MMR-based approach.}
	\label{fig:approach}
\end{figure}

A visual representation of our current base approach can be seen in Figure ~\ref{fig:approach}. The \textit{k} constant is the number of documents we take from the cluster for SDS generation. For datasets with only a few documents per cluster such as Multi-News, \textit{k} is usually all documents in the cluster. For datasets which contain 100 or more documents per cluster such as WCEP, \textit{k} is usually smaller than the number of documents in the cluster. \textit{m} is the desired output length of the final summary in sentences, ultimately defined as \textit{n}, the number of sentences in the model-generated summary with the highest MMR score, plus \textit{l}, the optimal number of additional sentences to have. \textit{l} can have a range of values, but we found it optimal to set \textit{l} to \textit{n} * \textit{p}, with \textit{p} being the percentage of final output sentences to be extracted using MMR in the \textit{m} * \textit{p} portion of the approach. The available sentences from each model's output is reduced subtracting the maximum number of MMR-removed sentences \textit{r} from the given model output's number of sentences \textit{d}. The MMR reduction is further explained in Section ~\ref{mmrreduction}.

 For example, this approach gives the option to specify a final output of length \textit{n + 2} and a \textit{p} value of 0.9, which would reconstitute 90\% of the final output using MMR, with the remainder of the \textit{n + 2} output being composed of unchanged output from the best-scoring model. For optimization in our case, we set \textit{l} to \textit{max(1, n*p)}, which means the entire best model output of \textit{n} length is unchanged, but these sentences are concatenated with at least 1 sentence which is extracted using MMR, and as many as \textit{n*p} sentences. Our use of a max expression ensures that the final output is never a simple repetition of the best-scoring model's output, as the intention of our research is to explore the improvement of model output using MMR.

After each model produces its SDS summary, as well as its MDS summary if applicable, each model's summary is split into sentences. By sorting each model's output sentences as whole sentences, we preserve the readability and coherence of the resulting summary without the need for further language processing. If the outputs were instead processed using individual words or phrases, the words would be rearranged to lose coherence without additional parameters added to this approach, which is beyond the scope of this research.

The SDS outputs for the given model are then concatenated into a single output from the model. However, this step of the process requires a limit to be placed on the number of summaries combined into the final SDS output in the case of a large imbalance between cluster size and document length in the dataset. For example, if the dataset has 100 documents per cluster and 1 sentence per document, then the concatenation of all SDS summaries would clearly result in an output too long to be comparable to the expected gold summary. Therefore, the number of SDS summaries returned per model for our approach is at maximum the average number of sentences per document. To extract the best given number of SDS outputs from all of the model's SDS outputs, we use the MMR algorithm with a focus on relevance.

\subsection{MMR Reduction}
\label{mmrreduction}
In order to reduce the chance of corrupt or irrelevant information being included in our final MMR calculation while increasing the relevance, we also make use of MMR to reduce the sentences of each model's output by a certain percentage if the model is not pretrained on the given dataset. Early experimentation revealed that this additional usage of MMR improves the results when there are a large number of sentences to select from, potentially revealing a minor shortcoming in the use of LDA topic word queries. However, this reduction is only performed in the case of datasets that produce numerous sentences per summary. When there are sufficient sentences per summary, each model's summary is reduced by MMR to a size of max(1, $n - $r), where \textit{d} is the size of the summary in sentences and \textit{r} is the number of sentences to reduce by. \textit{r} is defined by max(1, $d \cdot R$), with \textit{R} being the percentage to reduce by, which can be as small as 1\%. In our experiments, this reduction percentage is small enough that each summary is reduced by 1, although in datasets with more sentences a different optimal percentage might be found.

Preliminary research indicated that this simple reduction of sentences improves overall summarization performance. It was also found that some model outputs may benefit from the use of certain similarity measurement methods, such as the use of TF-IDF vectorization with cosine similarity for one model and Doc2Vec vectorization with cosine similarity for another. This use of MMR for model output reduction essentially selects the $(d-1)$ most-relevant sentences from each model's output, where \textit{d} is that number of sentences originally generated by the given model. In the case of datasets that produce summaries of only one or two sentences, no MMR reduction is used.

\subsection{MMR Query Document} For the query document in our implementation of the MMR formula, we use a small sequence of words generated via Latent Dirichlet Allocation (LDA) topic modeling. LDA is a statistical model which incorporates Bayesian calculations to infer the most probable topic words from the given corpus \cite{Blei03latentdirichlet}. We also use a low lambda constant in our MMR calculations to prioritize diversity among the input documents over relevance to the query document.

\subsection{Final Output} After applying this MMR calculation to reduce the model output if applicable, we combine all summaries into a single sequence. We then select the m best sentences from this single sequence using our MMR calculation, where m is the desired final number of sentences in our approach's output and the best sentences are those which best satisfy the diversity and relevance requirements  set by the lambda constant. This new sequence of the m best sentences is our final summary. We expect our MMR-based, diversity-focused combination approach to increase the final summary's similarity to human output without sacrificing the coherence and readability of the output already generated by the state-of-the-art models.

\section{Experiments}

We conduct two experiments to test our proposed approach: 1) The few-shot Multi-News experiment, for which one model was pretrained on the dataset, and 2) The zero-shot WCEP experiment, for which no models were pretrained on the dataset. These two experiments demonstrate the performance of our approach in few-shot and zero-shot applications, respectively. Additionally, the WCEP experiment demonstrates how our approach handles larger and potentially-unclean data. Our implementations for both experiments use the PyTorch numerical framework \cite{NEURIPS20199015}. We ran both of our experiments on Google's Colaboratory Pro platform using their GPUs \footnote{https://colab.research.google.com}. The GPUs available on the Colaboratory platform include Nvidia K80s, T4s, P4,s and P100s, although the exact GPU being used cannot be selected by the end user \cite{ColabFAQ}. In each experiment, we run an optimization library for the parameters of our approach, including the lambda constant determining relevance versus diversity, the percentage of the final output to be MMR-generated with a minimum of one sentence, the number of LDA topics to generate, and the number of words to generate per LDA topic. The LDA parameters determine the size and diversity of the query document to be used for the MMR calculation, with the number of topics primarily determining the diversity of information covered by the query document, and the number of words determining allowing the topic phrases to be expounded upon or reworded via a longer string of words. For both experiments, we compare our final MMR-generated summaries, as well as the intermediary model-generated summaries, with the human-generated summaries from the given dataset using ROUGE metrics \cite{lin-2004-rouge}. This set of metrics compares the co-occurrence of text sequences of various lengths to produce the precision and recall of the generated summaries, as well as the combined F-measure score. We use these metrics because they are sufficient approximations for summarization performance, particularly in experiments such as ours in which more expensive evaluations methods such as human evaluation are not available.

\subsection{Multi-News Experiment}
The Multi-News Experiment is a few-shot application, meaning some of the models have been pretrained on the dataset in question, and the system has some familiarity with the information to be processed. The purpose of this experiment is to determine the utility of this approach when a model is available which is pretrained on the dataset in question.

\subsection{Multi-News Dataset} For our corpus in this experiment, we use the Multi-News dataset, which consists of news article document clusters paired with professionally-written summaries, all of which are in the English language \cite{fabbri-etal-2019-multi}. There are 2 to 10 documents in each document cluster in this dataset, with each cluster containing an average of 82.73 sentences \cite{fabbri-etal-2019-multi}. The 5622 multi-document clusters from the testing subset of this dataset are split for single-document pretrained models using the dataset's special separation token.

\subsection{Models Used for Multi-News} For the models used to generate our summaries for combination, we employ three extractive models and five abstractive models. One of the extractive models is the matching model used by Zhong et al. \cite{zhong-etal-2020-extractive} in their implementation of summarization as semantic text-matching, known as MatchSum. This extractive model is pretrained on the Multi-News dataset, and produces state-of-the-art results. The other two extractive models were the XLNet\textsubscript{base} model \cite{yang2020xlnet} and the GPT-2\textsubscript{base} model \cite{Radford2019LanguageMA}, for which we use the model versions available through the Bert Extractive Summarizer library \footnote{https://pypi.org/project/bert-extractive-summarizer/}. This library was developed for an extractive summarization approach for lectures \cite{miller2019leveraging}. The models available through this library are simply placed within the architecture with no need for finetuning \cite{miller2019leveraging}. Multiple models are available through this library, although they were not included in the library's seminal research paper.

The five abstractive models we use for summarization are: the PEGASUS\textsubscript{multi-news} model \cite{zhang2020pegasus}, the BART\textsubscript{large-cnn} model \cite{lewis-etal-2020-bart}, the T5\textsubscript{large} model \cite{raffel2020exploring}, the ProphetNet\textsubscript{large-cnndm} model \cite{qi-etal-2020-prophetnet}, and the LED model \cite{beltagy2020longformer}. For simple implementation of these models, we employ the HuggingFace Transformers library where possible \cite{wolf2020huggingfaces}. For models which are not trained on the Multi-News dataset, we either use the large base pretrained model as in the case of T5, the base pretrained model as in the case of GPT-2, XLNet and LED, or the models pretrained on the CNN-DM dataset \cite{hermann2015teaching,nallapati-etal-2016-abstractive-alt} as in the case of BART and ProphetNet. In accordance with our proposed approach, due to the sufficient number of sentences in the Multi-News dataset, the outputs of the models which are not pretrained on the Multi-News dataset are reduced to a maximum of 1 and the model output length minus the MMR-reduction constant.

\subsection{Multi-News Experiment Parameters} We considered the use of two similarity measures for our Multi-News experiment: our TF-IDF similarity measure, which combines cosine similarity with a TF-IDF vectorizer, and our Doc2Vec similarity measure, which combines cosine similarity with a Doc2Vec vectorizer. More specifically, we use the Distributed Bag-of-Words (DBOW) model implementation within Gensim’s Doc2Vec architecture \cite{le2014distributed}, which uses predictions from randomly-sampled words in the document. We did not alter the seed of the vectorizer's random sampling.

Because we did not need to use single documents from the clusters, and because WMD was found to be inferior for the tasks of both MMR reduction and the final MMR combination, we did not use the WMD similarity measure in this experiment. Preliminary testing revealed that the vectorization method significantly affects model output selection, particularly with LED model outputs. For the final MMR combination method, we found the Doc2Vec similarity measure to be most effective. This method was also found to be most effective for the MMR-reduction of all model outputs with the exception of LED, for which we used the TF-IDF similarity measure.

Our MMR parameters were consistent across all usages, and were optimized using the black box optimization library Scikit-Optimization \footnote{https://scikit-optimize.github.io/stable/}. The optimal lambda constants, MMR percentages, LDA topic amounts, and LDA word amounts can be seen in Table~\ref{tab:mnews-optimized-parameters}

\begin{table}[h]
\centering
\begin{tabular*}{\linewidth}{l|c|c}
\hline
\textbf{Parameter} & \textbf{TF-IDF value} & \textbf{Doc2Vec value}\\
\hline
lambda constant & 0.975 & 0.808\\
MMR percentage & 0.107 & 0.106 \\
LDA topics & 3 & 5\\
Words per topic & 7 & 6\\
\hline
\end{tabular*}
\caption{Optimized MMR combination and LDA parameters for Multi-News.}
\label{tab:mnews-optimized-parameters}
\end{table}

\subsection{WCEP Experiment}
The WCEP Experiment is a zero-shot application, meaning none of the models are pretrained on the given dataset, and the system as a whole has no familiarity with the data to be processed. The purpose of this experiment is to determine the usefulness of this approach when no model is available which is pretrained on the given dataset.

\subsection{WCEP Dataset} For this experiment, we use the Wikipedia Current Events Portal (WCEP) dataset, which consists of short, human-written summaries of news events, the articles for which are all extracted from the Wikipedia Current Events Portal \cite{gholipour-ghalandari-etal-2020-large}. Each document cluster contains a large quantity of automatically-extracted articles. There are two primary versions of this dataset: the full version (WCEP-total) and the truncated version (WCEP-100). The full version consists of 2.39 million articles, while the truncated version consists of 650,000 articles. Also, each article cluster is limited to 100 articles in the truncated version, while the full version can contain as many as 8411 articles per cluster. The test split of the WCEP-100 dataset consists of 1,022 article clusters, of which we perform our experiment on 500 due to cost limitations. We use the WCEP-100 dataset version available through the WCEP dataset authors' GitHub repository \footnote{https://github.com/complementizer/wcep-mds-dataset}.

\subsection{Special Considerations} The large number of documents per cluster posed two major differences from our Multi-News experiment. First, it would be prohibitively expensive for each model to generate SDS output for each of the 100 documents in each cluster. Therefore, our k constant from our proposed approach is more relevant in this experiment, as our database clusters contain more than the desired number of documents. Only the 10 most relevant documents are summarized from each cluster, which are selected using MMR.

Our selection method also reduces the number of irrelevant, overly brief, unintelligible, or otherwise bad data selected from the large data clusters of the dataset. Second, each model is generating SDS outputs only, and there are more documents per cluster than sentences per document even after our MMR selection. Therefore, in accordance with our method previously described, the summary selection  method rather than the concatenation method is used to construct the best model output to append our MMR-extracted sentences to. This use of summary selection means that each model's output is represented by 10 smaller summaries rather than 1 larger summary which would be incomparable to the much smaller expected summaries of the dataset. The best summary of the best model is selected using MMR with a focus on relevance.

Additionally, we faced the problem of which of the SDS outputs of the best model to choose to append our MMR-selected output to. We wanted to select the best summary from the best-performing model without apriori evaluation, and with a suitable final number of sentences for comparison with the summaries of the sentences themselves. To select the best summary, we uses the cosine similarity measure with the Doc2Vec vectorizer, which is equivalent to the MMR algorithm with an output size of 1 sentence. We considered appending the MMR output to each of the best model's single-document summaries prior to selecting the best one. While this method did perform better than the baseline models, it was not optimal, and increased the likelihood of an unreadable final summary being generated. We therefore chose to append the MMR output only after the best summary to append to was selected using our Doc2Vec similarity measure.

Also, because the documents in the WCEP often contain only one or two sentences, and because the summaries generated from it are similarly brief, the use of MMR to reduce each model's output would be redundant. Therefore, no MMR reduction is performed in this experiment, in accordance with the procedures of our approach.

\subsection{Models Used for WCEP} For this experiment, we used 7 abstractive models and 3 extractive models. The model selection was slightly different in this experiment due to the different dataset. The MatchSum extractive model used in our Multi-News experiment was not available for this dataset, so we instead included both extractive and abstractive implementations of the XLNet, BART, and GPT-2 models. Additionally, we included the abstractive implementations of the ProphetNet, GPT-2, LED, T5, XLNet, PEGASUS, and BART models. As was the case with our Multi-News experiment, all of the abstractive implementations used were taken from the HuggingFace platform \footnote{https://huggingface.co/} created by Wolf et al. \cite{wolf2020huggingfaces}.  As there was no MatchSum model in this experiment, all of the extractive implementations we used were taken from the Bert Extractive Summarizer library \footnote{https://pypi.org/project/bert-extractive-summarizer/} developed by Miller \cite{miller2019leveraging}. We ensured that the model versions used in the extractive implementations were the same as the model versions used in the abstractive implementations.

\subsection{WCEP Experiment Parameters} As with the Multi-News Experiment, we consider the use of our TF-IDF, Doc2Vec, and WMD similarity measures for this experiment. As in the Multi-News experiment, we used Gensim's DBOW Doc2Vec vectorizer with the default random sampling seed. Given the added need for similarity-based selection in this experiment with a larger dataset and smaller summaries, we consider the three uses of similarity in this implementation: Sim\textsubscript{0}, Sim\textsubscript{1}, and Sim\textsubscript{2}. Sim\textsubscript{0} is the similarity measure used to select the best summary from the best model to append our MMR output to. Sim\textsubscript{1} and Sim\textsubscript{2} are the similarity measures used in the query relevance and diversity calculations of the MMR algorithm, respectively. Note that the Sim\textsubscript{2} similarity measure is rarely used in this implementation, because of the small number of returned sentences as well as the high lambda constant and relevance prioritization. Our WCEP experiment parameters were optimized using the black box optimization library Optuna \footnote{https://optuna.org/}. The optimized similarity measures we used, as well as the optimal MMR and LDA parameters, are included in Table~\ref{tab:wcep-optimized-parameters}.

\begin{table}[h]
\centering
\begin{tabular*}{\linewidth}{l|c}
\hline
\textbf{Parameter} & \textbf{Value}\\
\hline
lambda constant & 0.997 \\
MMR percentage & 0.298 \\
LDA topics & 5 \\
LDA words per topic & 2 \\
Sim\textsubscript{0} & Doc2Vec \\
Sim\textsubscript{1} & WMD \\
Sim\textsubscript{2} & TF-IDF \\
\hline
\end{tabular*}
\caption{Optimized WCEP MMR combination and LDA parameters.}
\label{tab:wcep-optimized-parameters}
\end{table}

\section{Results}

The resulting F-measure ROUGE scores of both experiments for our approach can be seen in Table~\ref{tab:results}. Because the Multi-News Experiment is a few-shot application rather than a zero-shot application, a comparison with the Mutli-News results of an additional pretrained model, MatchSum, is included.

\begin{table}[ht]
\centering
\begin{tabular*}{\linewidth}{l|c|c|c}
\hline
\multicolumn{4}{c}{\textbf{Multi-News}} \\
\hline
\textbf{Model} & \textbf{R1} & \textbf{R2} & \textbf{RL}\\
\hline
ProphetNet\textsubscript{large-cnndm} & 32.01 & 10.43 & 16.46	\\
GPT-2\textsubscript{base} & 32.56 & 9.87 & 16.34			\\
XLNet\textsubscript{base} & 32.76 & 9.97 & 16.38			\\
T5\textsubscript{large} & 32.95 & 10.27 & 16.62			\\
LED\textsubscript{base} & 33.27 & 10.83 & 16.84			\\
BART\textsubscript{large-cnn} & 36.64 & 12.09 & 18.19		\\
MatchSum\textsubscript{BERT-base, multi-news} & 43.98 & 15.91 & 20.94	\\
PEGASUS\textsubscript{multi-news}(SDS) & 40.25 & 16.26 & 19.43\\
PEGASUS\textsubscript{multi-news}(MDS) & 45.79 & \textbf{18.48} & \textbf{24.27}\\
MMR-combination (ours) & \textbf{46.23} & 18.30 & 21.25			\\
\hline
\multicolumn{4}{c}{\textbf{WCEP}} \\
\hline
\textbf{Model} & \textbf{R1} & \textbf{R2} & \textbf{RL}\\
\hline
XLNet\textsubscript{base} & 12.89 & 2.70 & 9.48			\\
ProphetNet\textsubscript{large-cnndm} & 14.19 & 2.36 & 10.56	\\
GPT-2\textsubscript{small} & 15.89 & 3.08 & 11.18			\\
T5\textsubscript{large} & 24.14 & 5.98 & 16.95			\\
LED\textsubscript{base} & 24.38 & 8.55 & 16.72			\\
BART\textsubscript{large-cnn}(EXT) & 27.17 & 9.39 & 19.39	\\
PEGASUS\textsubscript{xsum} & 27.33 & 8.38 & 19.19		\\
GPT-2\textsubscript{small}(EXT) & 27.33 & 9.42 & 19.52		\\
XLNet\textsubscript{base}(EXT) & 27.41 & \textbf{9.49} & 19.48	\\
BART\textsubscript{large-cnn} & 28.12 & 8.84 & 19.71		\\
MMR-combination (ours) & \textbf{30.74} & 9.18 & \textbf{21.57}\\
\hline
\end{tabular*}
\caption{Resulting F1-scores of ROUGE-1 (R1), ROUGE-2 (R2), and ROUGE-L (RL) metrics from test sets of Multi-News and WCEP datasets. (SDS) denotes that the model was used to summarize the individual documents from the document cluster, and (MDS) denotes that the model was used to summarize the complete document cluster as a single flattened document.}
\label{tab:results}
\end{table}

As shown in Table 1, we also observe that the ROUGE-1 score of our MMR-based combination approach is higher than all of the models used in both experiments. However, in the Multi-News experiment, the ROUGE-2 and ROUGE-L scores of our approach are lower than those of the MDS PEGASUS model implementation. While the ROUGE-2 and ROUGE-L scores of our MMR-based approach are lower than those of the PEGASUS model, they remain higher than those of all other models with the exception of PEGASUS. This relative discrepancy in ROUGE scores seems to indicate that, while our approach does not focus on the properties of multiple-word strings within the summaries, it nevertheless preserves state-of-the-art performance in n-gram inclusion.

Due to the increased ROUGE-1 score, it seems possible that there is at least a limited direct correlation between term relevance and ROUGE scores for the length of the terms being related to the LDA query document. Furthermore, these scores represent the use of only two models which are pretrained on the correct dataset, the combination of which was determined via preliminary experimentation to decrease all ROUGE scores relative to PEGASUS. We can therefore infer that the increase in ROUGE-1 score is due to the inclusion of models which are not pretrained on the Multi-News dataset. If these models were pretrained on the Multi-News dataset, we find it possible that the ROUGE-2 and ROUGE-L scores would not decrease as much relative to the leading pretrained model.

We see an additional increase in the ROUGE-L score in the zero-shot WCEP experiment, in which no models were pretrained on the given dataset. We also observe that the relative improvement of our approach over state-of-the-art methods is larger in the zero-shot WCEP experiment than in the few-shot Multi-News experiment, indicating that our approach is particularly effective in fewer-shot and zero-shot MDS applications. Due to our method of extracting whole sentences rather than words and phrases, readability and coherence is largely preserved.

These results indicate that our approach is useful in fewer-shot and zero-shot applications, such as applications in which the available data is too scarce or the cost of finetuning is too high. Furthermore, these results suggest the possibility that our approach grows more useful as the availability of finetuned models decreases, although this possibility would need to be more thoroughly tested to confirm.

\section{Conclusion}
We present a novel MMR-based framework that improves upon state-of-the-art pretrained models' performances for the task of MDS, particularly in fewer-shot and zero-shot applications where the cost for finetuning is too high or the available training data is too scarce. This framework increases single-term similarity to human-generated summaries by increasing the single-term relevance of the summaries generated by pretrained models, and further combining the summaries using MMR. These MMR-generated summaries maintain state-of-the-art quality and readability in addition to the improved single-term similarity results. Our approach serves to further demonstrate the potential for MMR as an effective tool for the task of automatic text summarization. We also demonstrate the potential for diversity among summary words to improve aspects of summary quality. It is unclear whether an MMR-based approach which prioritized additional n-gram relevance as opposed to only single-term relevance would produce superior results, and this uncertainty could be the basis for future research.

\section{Bibliographical References}\label{reference}


\begin{thebibliography}{}

\bibitem[\protect\citename{Beltagy \bgroup et al.\egroup
  }2020]{beltagy2020longformer}
Beltagy, I., Peters, M.~E., and Cohan, A.
\newblock (2020).
\newblock Longformer: The long-document transformer.
\newblock In {\em arXiv}.

\bibitem[\protect\citename{Blei \bgroup et al.\egroup
  }2003]{Blei03latentdirichlet}
Blei, D.~M., Ng, A.~Y., Jordan, M.~I., and Lafferty, J.
\newblock (2003).
\newblock Latent dirichlet allocation.
\newblock {\em Journal of Machine Learning Research}, 3:2003.

\bibitem[\protect\citename{Carbonell and Goldstein}1998]{101145/290941291025}
Carbonell, J. and Goldstein, J.
\newblock (1998).
\newblock The use of mmr, diversity-based reranking for reordering documents
  and producing summaries.
\newblock In {\em Proceedings of the 21st Annual International ACM SIGIR
  Conference on Research and Development in Information Retrieval}, SIGIR '98,
  page 335–336, New York, NY, USA. Association for Computing Machinery.

\bibitem[\protect\citename{Cohan \bgroup et al.\egroup
  }2018]{cohan-etal-2018-discourse}
Cohan, A., Dernoncourt, F., Kim, D.~S., Bui, T., Kim, S., Chang, W., and
  Goharian, N.
\newblock (2018).
\newblock A discourse-aware attention model for abstractive summarization of
  long documents.
\newblock In {\em Proceedings of the 2018 Conference of the North {A}merican
  Chapter of the Association for Computational Linguistics: Human Language
  Technologies, Volume 2 (Short Papers)}, pages 615--621, New Orleans,
  Louisiana, June. Association for Computational Linguistics.

\bibitem[\protect\citename{Daiya and Singh}2018]{daiya-singh-2018-using}
Daiya, D. and Singh, A.
\newblock (2018).
\newblock Using statistical and semantic models for multi-document
  summarization.
\newblock In {\em Proceedings of the 30th Conference on Computational
  Linguistics and Speech Processing ({ROCLING} 2018)}, pages 169--183, Hsinchu,
  Taiwan, October. The Association for Computational Linguistics and Chinese
  Language Processing (ACLCLP).

\bibitem[\protect\citename{Devlin \bgroup et al.\egroup
  }2019]{devlin-etal-2019-bert}
Devlin, J., Chang, M.-W., Lee, K., and Toutanova, K.
\newblock (2019).
\newblock {BERT}: Pre-training of deep bidirectional transformers for language
  understanding.
\newblock In {\em Proceedings of the 2019 Conference of the North {A}merican
  Chapter of the Association for Computational Linguistics: Human Language
  Technologies, Volume 1 (Long and Short Papers)}, pages 4171--4186,
  Minneapolis, Minnesota, June. Association for Computational Linguistics.

\bibitem[\protect\citename{Dong \bgroup et al.\egroup }2019]{dong2019unified}
Dong, L., Yang, N., Wang, W., Wei, F., Liu, X., Wang, Y., Gao, J., Zhou, M.,
  and Hon, H.-W.
\newblock (2019).
\newblock Unified language model pre-training for natural language
  understanding and generation.
\newblock In {\em arXiv}.

\bibitem[\protect\citename{Fabbri \bgroup et al.\egroup
  }2019]{fabbri-etal-2019-multi}
Fabbri, A., Li, I., She, T., Li, S., and Radev, D.
\newblock (2019).
\newblock Multi-news: A large-scale multi-document summarization dataset and
  abstractive hierarchical model.
\newblock In {\em Proceedings of the 57th Annual Meeting of the Association for
  Computational Linguistics}, pages 1074--1084, Florence, Italy, July.
  Association for Computational Linguistics.

\bibitem[\protect\citename{Gholipour~Ghalandari \bgroup et al.\egroup
  }2020]{gholipour-ghalandari-etal-2020-large}
Gholipour~Ghalandari, D., Hokamp, C., Pham, N.~T., Glover, J., and Ifrim, G.
\newblock (2020).
\newblock A large-scale multi-document summarization dataset from the
  {W}ikipedia current events portal.
\newblock In {\em Proceedings of the 58th Annual Meeting of the Association for
  Computational Linguistics}, pages 1302--1308, Online, July. Association for
  Computational Linguistics.

\bibitem[\protect\citename{Google}2021]{ColabFAQ}
Google.
\newblock (2021).
\newblock {Colaboratory} frequently asked questions.

\bibitem[\protect\citename{Hermann \bgroup et al.\egroup
  }2015]{hermann2015teaching}
Hermann, K.~M., Kočiský, T., Grefenstette, E., Espeholt, L., Kay, W.,
  Suleyman, M., and Blunsom, P.
\newblock (2015).
\newblock Teaching machines to read and comprehend.
\newblock In {\em arXiv}.

\bibitem[\protect\citename{Khatri \bgroup et al.\egroup
  }2018]{khatri2018abstractive}
Khatri, C., Singh, G., and Parikh, N.
\newblock (2018).
\newblock Abstractive and extractive text summarization using document context
  vector and recurrent neural networks.
\newblock In {\em arXiv}.

\bibitem[\protect\citename{Lan \bgroup et al.\egroup }2020]{lan2020albert}
Lan, Z., Chen, M., Goodman, S., Gimpel, K., Sharma, P., and Soricut, R.
\newblock (2020).
\newblock Albert: A lite bert for self-supervised learning of language
  representations.
\newblock In {\em arXiv}.

\bibitem[\protect\citename{Le and Mikolov}2014]{le2014distributed}
Le, Q.~V. and Mikolov, T.
\newblock (2014).
\newblock Distributed representations of sentences and documents.
\newblock In {\em arXiv}.

\bibitem[\protect\citename{Lee and Lee}2017]{lee-lee-2017-automatic}
Lee, G.~H. and Lee, K.~J.
\newblock (2017).
\newblock Automatic text summarization using reinforcement learning with
  embedding features.
\newblock In {\em Proceedings of the Eighth International Joint Conference on
  Natural Language Processing (Volume 2: Short Papers)}, pages 193--197,
  Taipei, Taiwan, November. Asian Federation of Natural Language Processing.

\bibitem[\protect\citename{Lewis \bgroup et al.\egroup
  }2020]{lewis-etal-2020-bart}
Lewis, M., Liu, Y., Goyal, N., Ghazvininejad, M., Mohamed, A., Levy, O.,
  Stoyanov, V., and Zettlemoyer, L.
\newblock (2020).
\newblock {BART}: Denoising sequence-to-sequence pre-training for natural
  language generation, translation, and comprehension.
\newblock In {\em Proceedings of the 58th Annual Meeting of the Association for
  Computational Linguistics}, pages 7871--7880, Online, July. Association for
  Computational Linguistics.

\bibitem[\protect\citename{Lin}2004]{lin-2004-rouge}
Lin, C.-Y.
\newblock (2004).
\newblock {ROUGE}: A package for automatic evaluation of summaries.
\newblock In {\em Text Summarization Branches Out}, pages 74--81, Barcelona,
  Spain, July. Association for Computational Linguistics.

\bibitem[\protect\citename{Liu \bgroup et al.\egroup }2019]{liu2019roberta}
Liu, Y., Ott, M., Goyal, N., Du, J., Joshi, M., Chen, D., Levy, O., Lewis, M.,
  Zettlemoyer, L., and Stoyanov, V.
\newblock (2019).
\newblock Roberta: A robustly optimized bert pretraining approach.
\newblock In {\em arXiv}.

\bibitem[\protect\citename{Mao \bgroup et al.\egroup
  }2020]{mao-etal-2020-multi}
Mao, Y., Qu, Y., Xie, Y., Ren, X., and Han, J.
\newblock (2020).
\newblock Multi-document summarization with maximal marginal relevance-guided
  reinforcement learning.
\newblock In {\em Proceedings of the 2020 Conference on Empirical Methods in
  Natural Language Processing (EMNLP)}, pages 1737--1751, Online, November.
  Association for Computational Linguistics.

\bibitem[\protect\citename{Miller}2019]{miller2019leveraging}
Miller, D.
\newblock (2019).
\newblock Leveraging bert for extractive text summarization on lectures.
\newblock In {\em arXiv}.

\bibitem[\protect\citename{Nallapati \bgroup et al.\egroup
  }2016]{nallapati-etal-2016-abstractive-alt}
Nallapati, R., Zhou, B., dos Santos, C., GuÌ‡l{\c{c}}ehre, {\c{C}}., and
  Xiang, B.
\newblock (2016).
\newblock Abstractive text summarization using sequence-to-sequence {RNN}s and
  beyond.
\newblock In {\em Proceedings of The 20th {SIGNLL} Conference on Computational
  Natural Language Learning}, pages 280--290, Berlin, Germany, August.
  Association for Computational Linguistics.

\bibitem[\protect\citename{Narayan \bgroup et al.\egroup
  }2018]{narayan-etal-2018-ranking}
Narayan, S., Cohen, S.~B., and Lapata, M.
\newblock (2018).
\newblock Ranking sentences for extractive summarization with reinforcement
  learning.
\newblock In {\em Proceedings of the 2018 Conference of the North {A}merican
  Chapter of the Association for Computational Linguistics: Human Language
  Technologies, Volume 1 (Long Papers)}, pages 1747--1759, New Orleans,
  Louisiana, June. Association for Computational Linguistics.

\bibitem[\protect\citename{Paszke \bgroup et al.\egroup }2019]{NEURIPS20199015}
Paszke, A., Gross, S., Massa, F., Lerer, A., Bradbury, J., Chanan, G., Killeen,
  T., Lin, Z., Gimelshein, N., Antiga, L., Desmaison, A., Kopf, A., Yang, E.,
  DeVito, Z., Raison, M., Tejani, A., Chilamkurthy, S., Steiner, B., Fang, L.,
  Bai, J., and Chintala, S.
\newblock (2019).
\newblock Pytorch: An imperative style, high-performance deep learning library.
\newblock In H.~Wallach, et~al., editors, {\em Advances in Neural Information
  Processing Systems 32}, pages 8024--8035. Curran Associates, Inc.

\bibitem[\protect\citename{Qi \bgroup et al.\egroup
  }2020]{qi-etal-2020-prophetnet}
Qi, W., Yan, Y., Gong, Y., Liu, D., Duan, N., Chen, J., Zhang, R., and Zhou, M.
\newblock (2020).
\newblock {P}rophet{N}et: Predicting future n-gram for
  sequence-to-{S}equence{P}re-training.
\newblock In {\em Findings of the Association for Computational Linguistics:
  EMNLP 2020}, pages 2401--2410, Online, November. Association for
  Computational Linguistics.

\bibitem[\protect\citename{Radford \bgroup et al.\egroup
  }2019]{Radford2019LanguageMA}
Radford, A., Wu, J., Child, R., Luan, D., Amodei, D., and Sutskever, I.
\newblock (2019).
\newblock Language models are unsupervised multitask learners.
\newblock In {\em Technical report}. OpenAI.

\bibitem[\protect\citename{Raffel \bgroup et al.\egroup
  }2020]{raffel2020exploring}
Raffel, C., Shazeer, N., Roberts, A., Lee, K., Narang, S., Matena, M., Zhou,
  Y., Li, W., and Liu, P.~J.
\newblock (2020).
\newblock Exploring the limits of transfer learning with a unified text-to-text
  transformer.
\newblock In {\em arXiv}.

\bibitem[\protect\citename{See \bgroup et al.\egroup }2017]{see-etal-2017-get}
See, A., Liu, P.~J., and Manning, C.~D.
\newblock (2017).
\newblock Get to the point: Summarization with pointer-generator networks.
\newblock In {\em Proceedings of the 55th Annual Meeting of the Association for
  Computational Linguistics (Volume 1: Long Papers)}, pages 1073--1083,
  Vancouver, Canada, July. Association for Computational Linguistics.

\bibitem[\protect\citename{Vaswani \bgroup et al.\egroup
  }2017]{vaswani2017attention}
Vaswani, A., Shazeer, N., Parmar, N., Uszkoreit, J., Jones, L., Gomez, A.~N.,
  Kaiser, L., and Polosukhin, I.
\newblock (2017).
\newblock Attention is all you need.
\newblock In {\em arXiv}.

\bibitem[\protect\citename{Wolf \bgroup et al.\egroup
  }2020]{wolf2020huggingfaces}
Wolf, T., Debut, L., Sanh, V., Chaumond, J., Delangue, C., Moi, A., Cistac, P.,
  Rault, T., Louf, R., Funtowicz, M., Davison, J., Shleifer, S., von Platen,
  P., Ma, C., Jernite, Y., Plu, J., Xu, C., Scao, T.~L., Gugger, S., Drame, M.,
  Lhoest, Q., and Rush, A.~M.
\newblock (2020).
\newblock Huggingface's transformers: State-of-the-art natural language
  processing.
\newblock In {\em arXiv}.

\bibitem[\protect\citename{Yang \bgroup et al.\egroup }2020a]{yang2020xlnet}
Yang, Z., Dai, Z., Yang, Y., Carbonell, J., Salakhutdinov, R., and Le, Q.~V.
\newblock (2020a).
\newblock Xlnet: Generalized autoregressive pretraining for language
  understanding.
\newblock In {\em arXiv}.

\bibitem[\protect\citename{Yang \bgroup et al.\egroup
  }2020b]{yang-etal-2020-ted}
Yang, Z., Zhu, C., Gmyr, R., Zeng, M., Huang, X., and Darve, E.
\newblock (2020b).
\newblock {TED}: A pretrained unsupervised summarization model with theme
  modeling and denoising.
\newblock In {\em Findings of the Association for Computational Linguistics:
  EMNLP 2020}, pages 1865--1874, Online, November. Association for
  Computational Linguistics.

\bibitem[\protect\citename{Zhang \bgroup et al.\egroup }2020]{zhang2020pegasus}
Zhang, J., Zhao, Y., Saleh, M., and Liu, P.~J.
\newblock (2020).
\newblock Pegasus: Pre-training with extracted gap-sentences for abstractive
  summarization.
\newblock In {\em arXiv}.

\bibitem[\protect\citename{Zhong \bgroup et al.\egroup
  }2020]{zhong-etal-2020-extractive}
Zhong, M., Liu, P., Chen, Y., Wang, D., Qiu, X., and Huang, X.
\newblock (2020).
\newblock Extractive summarization as text matching.
\newblock In {\em Proceedings of the 58th Annual Meeting of the Association for
  Computational Linguistics}, pages 6197--6208, Online, July. Association for
  Computational Linguistics.

\end{thebibliography}

\end{document}